\documentclass[3p]{elsarticle}

\usepackage{lineno,hyperref}
\usepackage{enumitem}
\usepackage{url}
\usepackage{graphicx} 
\usepackage{float}
\usepackage{amsmath}
\usepackage{booktabs}
\usepackage{subfig}
\usepackage{dsfont}
\usepackage{bm}
\DeclareMathOperator*{\argmax}{\arg\!\max}

\journal{Journal of Knowledge-Based Systems}

\begin{document}

\begin{frontmatter}

\title{A Post-processing Method for Detecting Unknown Intent of Dialogue System via Pre-trained Deep Neural Network Classifier} 

\author{Ting-En Lin}
\ead{lte17@mails.tsinghua.edu.cn}
\author{Hua Xu\corref{mycorrespondingauthor}}
\cortext[mycorrespondingauthor]{Corresponding author}
\ead{xuhua@tsinghua.edu.cn}

\address{State Key Laboratory of Intelligent Technology and Systems, Department of Computer Science and Technology, Tsinghua University, Beijing 100084, China \\
 Beijing National Research Center for Information Science and Technology(BNRist), Beijing 100084, China}

\begin{abstract}
With the maturity and popularity of dialogue systems, detecting user's unknown intent in dialogue systems has become an important task. It is also one of the most challenging tasks since we can hardly get examples, prior knowledge or the exact numbers of unknown intents. 
In this paper, we propose SofterMax and deep novelty detection (SMDN), a simple yet effective post-processing method for detecting unknown intent in dialogue systems based on pre-trained deep neural network classifiers. Our method can be flexibly applied on top of any classifiers trained in deep neural networks without changing the model architecture. We calibrate the confidence of the softmax outputs to compute the calibrated confidence score (i.e., SofterMax) and use it to calculate the decision boundary for unknown intent detection. Furthermore, we feed the feature representations learned by the deep neural networks into traditional novelty detection algorithm to detect unknown intents from different perspectives. Finally, we combine the methods above to perform the joint prediction. Our method classifies examples that differ from known intents as unknown and does not require any examples or prior knowledge of it. 
We have conducted extensive experiments on three benchmark dialogue datasets. The results show that our method can yield significant improvements compared with the state-of-the-art baselines\footnote{The code will be available at https://github.com/tnlin/SMDN}.

\end{abstract}

\begin{keyword}
Novelty detection \sep Open-world classification \sep Probability calibration \sep Platt scaling \sep Dialogue system \sep Deep neural network
\end{keyword}

\end{frontmatter}

\section{Introduction}
With the rapid development of natural language processing technologies, dialogue systems, such as virtual assistant and smart speaker, gradually plays an essential role in our daily life. The critical component of a dialogue system is to classify the user utterance into corresponding intents, which is usually from a predefined category of the training set. Then, the system can give the corresponding response based on the user intent. Nevertheless, the users' new intent appear regularly in real-world scenarios. It is important to identify the intents that have never appeared in the training set before. Although a small number of known intents can cover most conversations, the unknown intents in the remaining unsatisfied cases may have huge potential. For example, we can leverage unknown intents to identify business opportunities in task-oriented dialogue systems.

However, finding out user’s unknown intents is challenging \cite{lin-xu-2019-deep}. On the one hand, it is difficult to obtain prior knowledge \cite{lopez2015using} about unknown intents because there is usually a lack of examples of it. On the other hand, it is hard to estimate the exact number of unknown intents. Also, since user intents are strongly guided by prior knowledge and context, modeling high-level semantic concepts of intent are still problematic \cite{bi2018empirical, pota2019multilingual}. 

Only a few previous studies are related to unknown intent detection. For example, Kim and Kim \cite{Kim2018JointLO} try to train an intent classifier and out-of-domain detector jointly, but they still need out-of-domain examples during the training process. Generative methods \cite{Yu2017OpenCategoryCB} try to generate positive and negative examples from known classes by using adversarial learning to augment training data. However, the method does not work well in the discrete data space (like text data), and a recent study \cite{nalisnick2018do} suggests that this approach may not work well on real-world data. Brychc{\'i}n and Kr{\'a}l \cite{Brychcin2017UnsupervisedDA} model the through clustering. Still, it does not make good use of label information of known intents, and clustering results are usually unsatisfactory.

\begin{figure}
  \centering
  \includegraphics[width=0.5\linewidth ]{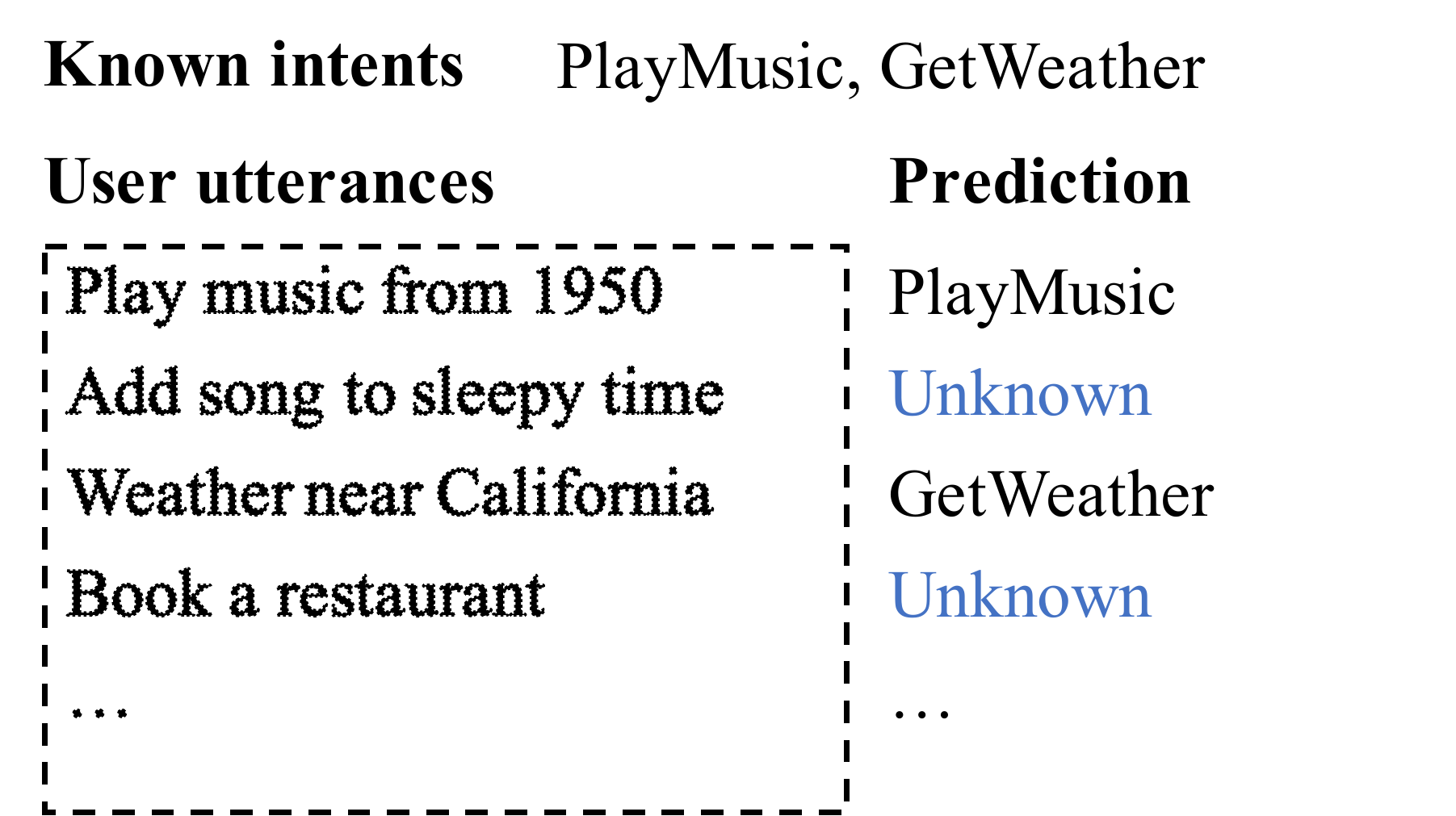}
  \caption{\label{example} An example of open-world classification. We use its concept as a surrogate for unknown intent detection.
   }
\end{figure}
How do we detect unknown intent without any prior knowledge about it? In \cite{Scheirer2013TowardOS, Fei2016BreakingTC}, a \emph{m}-class classifier should be able to reject examples from unknown class while performing \emph{m}-class classification tasks. It is because not all test classes are known in the training set, which forms a (\emph{m}+1)-class classification problem where the (\emph{m}+1)$^{th}$ class represents the unknown class. This problem is called open-world classification. We illustrate an example of open-world classification in Figure \ref{example}. As we can see, an open-world classifier should not only classify examples correctly but also reject examples which do not belong to any known classes. In this case, we can effectively use the label information of known intents to find unknown intents without any prior knowledge or examples of it and simplify the problem by grouping unknown intents into a single unknown class. The main idea is that if an example does not belong to any known intents, it will be considered as the unknown.

Inspired by open-world classification, we want to detect unknown user intents on top of existing classifiers. Nevertheless, traditional discriminative models (such as SVM) have limited ability to model the high-level semantic concept of intents. We can resolve this problem by using deep neural networks. In deep open-world classification, DOC \cite{Shu2017DOCDO} achieve the state-of-the-art performance by building a multi-class deep neural network classifier with a 1-vs-rest final layer of sigmoid functions and tightening the decision boundary of sigmoid function to reject unknown examples. However, it limits the activation function in the output layer to be sigmoid, whereas most classifiers use softmax in the output layer. 

To address the issues mentioned above, we propose SofterMax and deep novelty detection method (SMDN), a simple post-processing method that does not require any change to the model structure. To begin with, we calibrate the confidence of softmax outputs to get more reasonable probability distributions. Then, we calculate the per-class decision threshold with calibrated softmax output (i.e., SofterMax) to reject unknown examples. Furthermore, we combine traditional novelty detection algorithm with feature representations learned by deep neural networks to detect unknown intent. Last but not least, we scale the novelty score of the above methods into novelty probability and make the joint prediction with SofterMax. We conduct extensive experiments on three different dialogue datasets: SNIPS, ATIS, and SwDA. Then, we use macro F1-score as a metric to evaluate the effectiveness of the proposed method and compare the results with the state-of-the-art baselines. To the best of our knowledge, this is the first attempt to detect unknown intents in multi-turn dialogue systems without any hand-crafted features.

We summarize the main contributions of this paper as follows. Firstly, we proposed a flexible post-processing method for detecting unknown intent, which can be easily applied on top of any classifiers trained in deep neural networks without affecting the training process. Secondly, the proposed SofterMax method can learn a better decision threshold for detecting unknown intent through probability calibration. Thirdly, the proposed deep novelty detection method can combine traditional novelty detection algorithm with feature representations learned by the deep neural network. Then, we further make the joint prediction with SofterMax and deep novelty detection method. Finally, the experiments conducted on three benchmark dialogue datasets show the effectiveness and robustness of the proposed method comparing with the state-of-the-art baselines. 

The rest of this paper is structured as follows. In Section 2, we will briefly survey previous studies in open-world classification and others. In section 3, we will introduce the proposed method, SofterMax and Deep Novelty detection (SMDN), in details and report the experiment settings and results in Section 4. We will compare the results of different methods on three benchmark datasets with different settings in Section 5 and conclude the paper in Section 6.

\section{Related Work}
The research of detecting unknown class in dialogue systems is still in its infancy stage. In this section, we briefly review the related work in open-world classification, probability calibration, and novelty detection.

One-vs-Rest Support Vector Machine (1-vs-rest SVM) \cite{rifkin2004defense} is a margin-based method which trains the binary classifier for the individual classes. Given a class $C_i$ in a set of classes $C = \{{C_1, C_2, C_3, ... C_N}\}$, it treats the examples within $C_i$ as positive and the examples outside $C_i$ as negative. Then, we use those data to learn the decision boundary and generate N binary classifiers. We consider an example to be unknown if all binary classifiers classify it as negative. In the field of computer vision, 1-vs-set machine \cite{Scheirer2013TowardOS} is an extension of 1-vs-rest SVM. They first introduce the concept of open space risk as a measure of open-world classification. The main idea is that a classifier should not cover too much open space with few or no training data, thereby rejecting the unknown images. cbsSVM \cite{Fei2016BreakingTC} shares the similar ideas in text classification. However, these methods are all based on SVM, which fails to effectively capture the high-level semantic concept of intents comparing with deep neural networks.

Bendale and Boult \cite{Bendale2016TowardsOS} propose OpenMax, which extends the idea of reducing open space risk to the deep neural network. It suggests that a classifier should reduce the open risk in feature space rather than pixel space. They reject unknown images by utilizing the logits trained on deep neural networks. However, this method requires examples from unknown classes to tune the hyperparameters. It has been shown that DOC \cite{Shu2017DOCDO} outperforms all previous methods. We can apply DOC on top of deep neural network by using a 1-vs-rest final layer of sigmoids to preserve probability of each class. Furthermore, DOC tightens the decision boundary of sigmoid by calculating the probability threshold of each class through the statistical approach. This method tries to reduce open space risk in probability space. Yet, it limits the activation function of the output layer to be sigmoid.

Our work is also related to probability calibration. Guo et al. \cite{Guo2017OnCO} have shown that modern neural network tends to be overconfident compared to its ground truth correctness likelihood. Liang et al. \cite{liang2018enhancing} have shown that a well-calibrated model can benefit the outlier detection task, but it also requires examples from the unknown class to tune the hyperparameters. 

Novelty detection method allows us to detect abnormal examples (unknown class) that have not appeared in the training set during the testing stage. Local outlier factor (LOF) \cite{breunig2000lof} is a density-based novelty detection method. Its main idea is that if the local density of an example is significantly lower than its neighbors, it may be considered as abnormal. \cite{Sommer2017ADL} and \cite{prakhya2017open} use novelty detection algorithms to detect abnormal examples in the feature space extracted by autoencoder, and we can further improve this approach by learning better feature representations in deep neural networks.

\section{Proposed Method}
\begin{figure}[t!]
  \centering
  \includegraphics[width=0.99\linewidth]{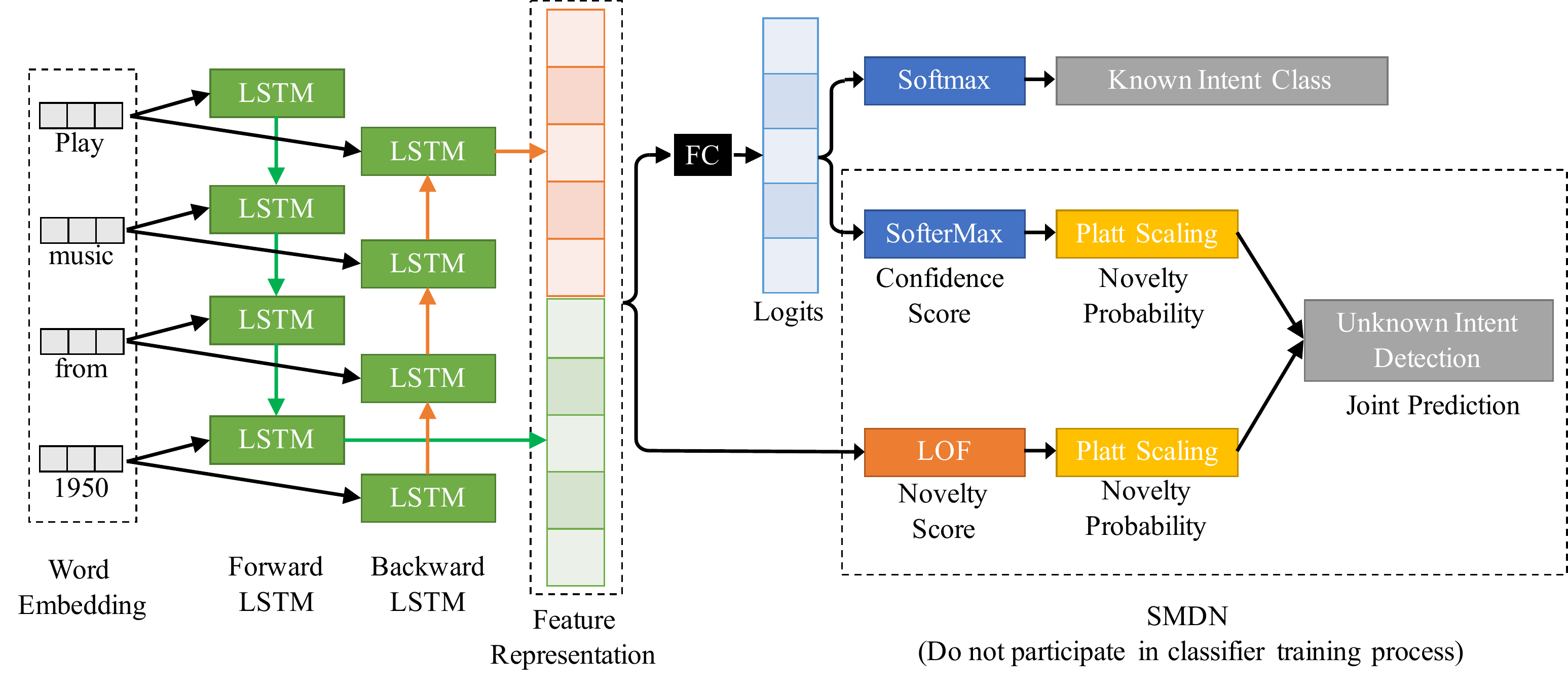}
  \caption{The experiment flowchart for unknown intent detection in the single-turn dialogue system. We use bidirectional LSTM to learn the feature representation of the intent. Then, we feed feature representation to fully-connected (FC) layer to obtain logits. Within the dashed box is the proposed SMDN method. We first train and validate the classifier on known samples, then test it with samples mixed with known and unknown intents.
    \label{model-single-figure}}
\end{figure}

In this section, we will give a detailed description of the proposed SofterMax and deep novelty detection method. The experiment flowchart for unknown intent detection in the single-turn and multi-turn dialogue system are shown in Figure \ref{model-single-figure} and \ref{model-multi-figure}, respectively. 

To start with, we train an intent classifier based on deep neural networks and take it as the prerequisite. Next, we calibrate the predicted confidence of the classifier through temperature scaling and then tighten the decision boundary of calibrated softmax (SofterMax) for detecting unknown intent. Also, we feed the feature representations learned by the deep neural network to the density-based novelty detection method, LOF, to discover unknown intents from different perspectives. Finally, we transform the confidence score of SofterMax and novelty score of LOF into novelty probability through Platt scaling \cite{platt1999probabilistic} and make the joint prediction.

\subsection{Classifiers}
The key idea of the proposed method is to detect unknown intents based on the existing intent classifier without modifying any network architecture. If an example is different from all known intents, it will be classified as the unknown. Because we detect the unknown intent based on the classifier, the performance of the intent classifier is crucial. The better the classifier we have, the better the result we get. Therefore, we implement classifiers similar to the state-of-the-art models of each dataset and compare different detection methods under the same classifier. 

\subsubsection{BiLSTM}
For modeling intents of single-turn dialogue dataset, SNIPS and ATIS, we use simplified structure described in \cite{Goo2018SlotGatedMF} and \cite{Wang2018ABB}. We use bidirectional LSTM (BiLSTM) \cite{Graves05bidirectionallstm} followed by a fully-connected layer and activation layer. BiLSTM is a many-to-one design where the dropout applies to the non-recurrent connections. We illustrate the experiment flowchart in Figure \ref{model-single-figure}. Given an utterance with maximum word sequence length $\ell $, we transform a sequence of input words $w_{1:\ell}$ into a sequence of m-dimensional word embedding $x_{1:\ell}$, which is used by forward and backward LSTM to produce feature representations $h$: 
\begin{align}
\overrightarrow{h_t} &= LSTM(x_t,\overrightarrow{c_{t-1}}) \\
\overleftarrow{h_t} &= LSTM(x_t,\overleftarrow{c_{t+1}}) \\
h &=[\overrightarrow{h_{\ell}};\overleftarrow{h_{1}}] ,\quad  z = Wh + b 
\end{align} 
where $x_t \in \mathds R^{m}$  denotes the m-dimension word embedding of input at time step $t$. $\overrightarrow{h_t}$ and $\overleftarrow{h_t}$ are the output vector of forward and backward LSTM, respectively.  $\overrightarrow{c_t}$ and $\overleftarrow{c_t}$ are the cell state vector of forward and backward LSTM, respectively. Before going through the softmax activation function, logit $z$ is the output of the fully connected layer, where the number of neurons is equal to the number of known classes. We consider the last output vector of forward LSTM ${\overrightarrow{h_{\ell}}}$ and the first output vector of backward LSTM $\overleftarrow{h_1}$ as the sentence representation learned by BiLSTM and concatenate them into $h$ to represent high-level semantic concepts learned by the model. We will take $h$ as the input of LOF and the $z$ as the input of SofterMax. 

\begin{figure}[t]
  \centering
  \includegraphics[width=0.99\linewidth]{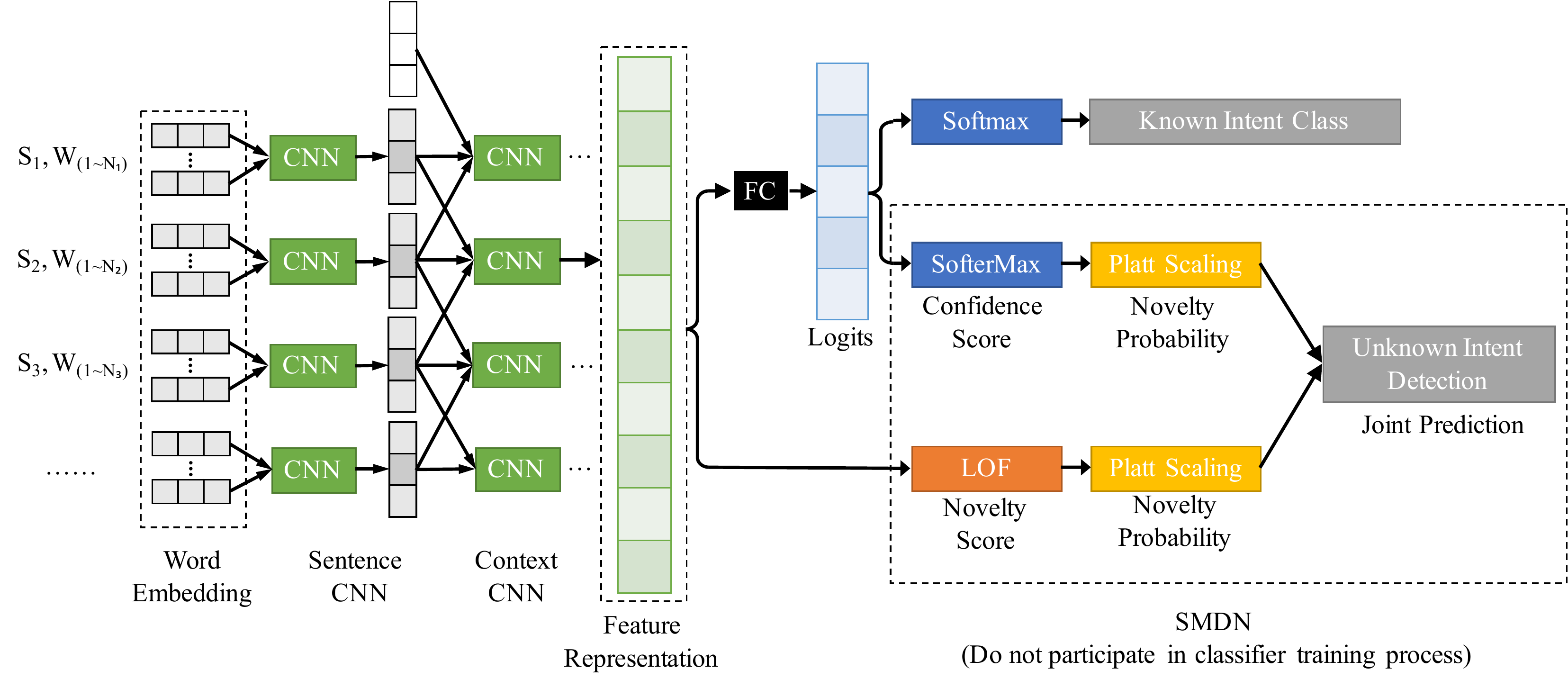}
  \caption{The experiment flowchart for unknown intent detection in the multi-turn dialogue system. We use hierarchical CNN to learn contextual feature representation of the intent. Then, we feed feature representation to fully-connected (FC) layer to obtain logits. Within the dashed box is the proposed SMDN method. We first train and validate the classifier on known samples, then test it with samples mixed with known and unknown intents. \label{model-multi-figure}}
\end{figure}

\subsubsection{CNN+CNN}
For modeling intents of multi-turn dialogue dataset, SwDA, we use hierarchical CNN (CNN+CNN) similar to the structure described in \cite{Liu2017UsingCI}. We illustrate the experiment flowchart in Figure \ref{model-multi-figure}.
Given a sentence in the conversation, we set the maximum length of the word sequences to $\ell$ and the context window size of sentence sequences to $c$. To begin with, we perform convolution operations on word sequences to produce features as the following: 
\begin{align}
	s_t = \text{ReLU}(W_{f1} \bullet x_{t:t+n-1} + b_{f1}) 
\end{align}
where $x_t \in \mathds R^{m}$ denotes the $m$-dimensional word embedding corresponding to the $t^{th}$ word in the sentence. ReLU is a non-linear function and $b_{f1}$ is a bias term. The symbol $\bullet$ denotes element-wise multiplication. The filter $W_{f1} \in \mathds R^{n \times m}$ performs convolution operation on a window of $n$ consecutive words to produce the feature $s_t$. Then, we apply the filter $W_{f1}$ on all possible window of words $\{x_{1:n}, x_{2:n+1}, \dots, x_{\ell_1-n+1:\ell_1}\}$ in the sentence to produce feature maps:
\begin{align}
	s = [s_1, s_2, \dots, s_{\ell_1-n+1}]
\end{align}
where $s \in \mathds R^{\ell_1-n+1} $. To get the specific feature $\hat s$ for the filter $W_{f1}$, we apply max-pooling operation over feature maps:
\begin{align}
	\hat s = \max\{s\}
\end{align}
where $\hat s$ is the scalar feature learned by $W_{f1}$. By repeating  convolution operations with $k_1$ different filters, we get the sentence representations $\textbf{z}$:
\begin{align}
	\textbf{z} = [\hat s_1, \dots, \hat s_{k_1}]
\end{align}
where $\textbf{z} \in \mathds R^{k_1} $ denotes the $k_1$-dimensional sentences vector.

Furthermore, we perform another convolution operations on a window of $c$ sentences to produce context representations of the target sentence as the following: 
\begin{align}
Z &= [\textbf{z}_{t-c-1},\dots, \textbf{z}_{t-1}, \textbf{z}_{t}, \textbf{z}_{t+1}, \dots, \textbf{z}_{t+c-1}] \\
	h_t &= \text{ReLU}(W_{f2} \bullet Z_{t:t+n-1} + b_{f2})
\end{align}
where $Z \in \mathds R^{2c-1 \times k_1} $ denotes the sentences representations within the context window size $c$ for the $t^{th}$ sentence in the conversation. The filter $W_{f2} \in \mathds R^{n \times k_1}$ performs convolution operation on a window of $n$ consecutive sentences to produce the feature $h_t$. Then, we apply the filter $W_{f2}$ on all possible window of sentences to produce feature maps:
\begin{align}
	h = [h_{t-c-1},\dots, h_{t-1}, h_{t}, h_{t+1}, \dots, h_{t+c-1}]
\end{align}

where $h \in \mathds R^{2c-1} $. To get the specific feature $\hat h$ for the filter $W_{f2}$, we apply max-pooling operation over feature maps:
\begin{align}
	\hat h = \max\{h\}
\end{align}
where $\hat h$ is the scalar feature learned by $W_{f2}$. Finally, we repeat  convolution operations with $k_2$ different filters to get the context representations $\textbf{r}$:
\begin{align}
	\textbf{r} = [\hat h_1, \dots, \hat h_{k_2}], \quad  z = W\textbf{r} + b
\end{align}
where $\textbf{r} \in \mathds R^{k_2} $ denotes the $k_2$-dimensional context vector of the target sentence. Logits $z$ is the output of the fully connected layer before passing the softmax activation function, where the number of neurons is equal to the number of known classes. Similar to the approach mentioned in BiLSTM model, we take \textbf{r} as the input of LOF and the $z$ as the input of SofterMax. As we can see in Figure \ref{model-single-figure} and \ref{model-multi-figure}, the proposed method can be flexibly applied to all kinds of pre-trained deep neural network classifiers.

\begin{figure}[t]
  \centering
  \includegraphics[width=0.5\linewidth ]{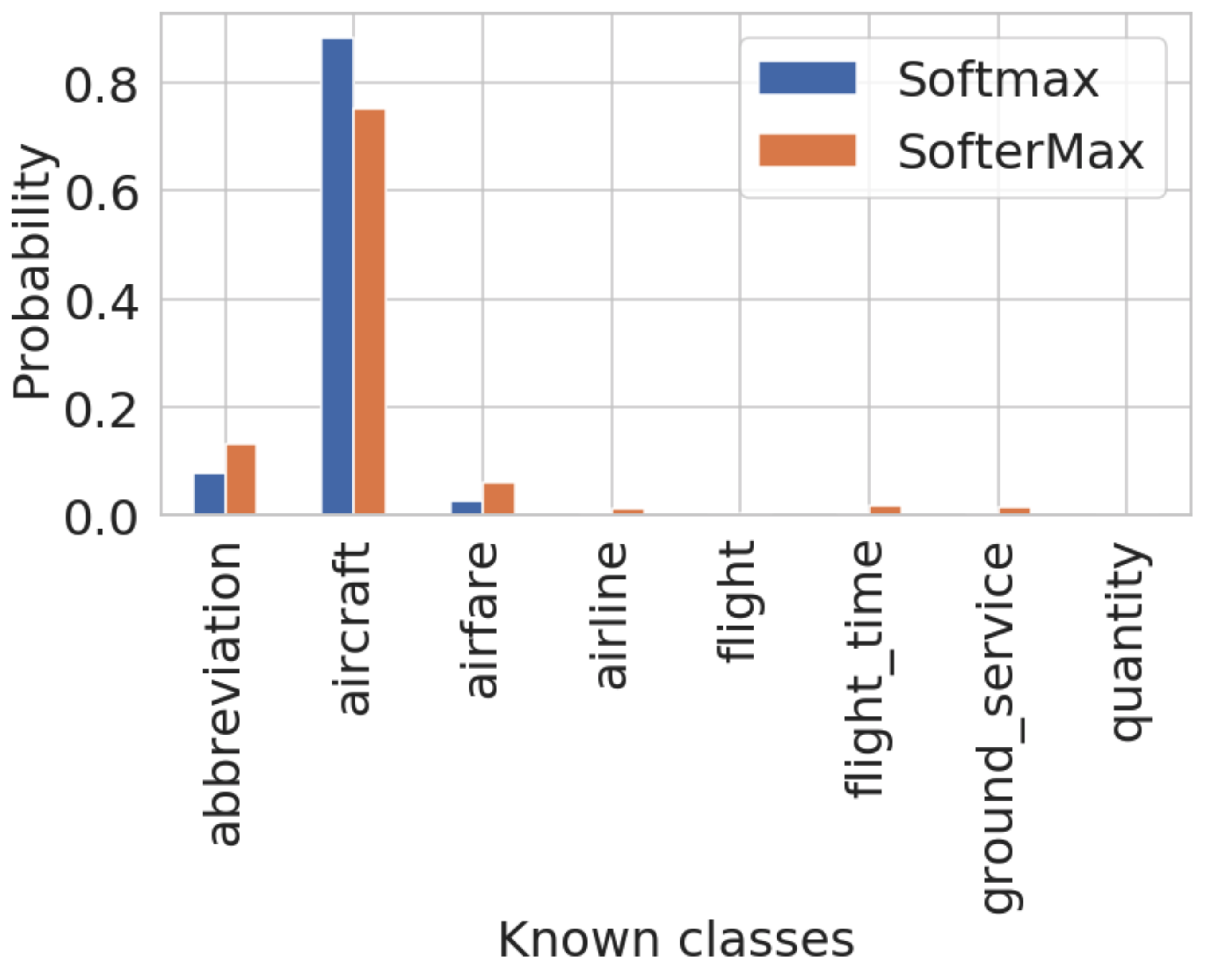}
  \caption{\label{softermax} An example illustrates the difference between Softmax and SofterMax with temperature parameter around 1.4. When T$>$1, SofterMax will output more conservative probability distribution over classes compared with Softmax. }
\end{figure}

\subsection{SofterMax}
Based on the pre-trained classifier, we calibrate the confidence of the softmax outputs to get more reasonable probability distributions. Then, we tighten the decision boundary of calibrated softmax (i.e., SofterMax) to reject unknown examples. In Figure \ref{softermax}, we can see the difference between softmax and SofterMax.

DOC \cite{Shu2017DOCDO} have shown that it is feasible to reject examples for unknown classes by reducing the open space risk in probability space. Meanwhile, the softmax output probabilities of deep neural network classifiers may be overconfident. It not only exposes too much open space risk (i.e., misclassified an example belonged to unknown class into a known class with high confidence) but also fails to provide a reasonable probability representation for output classes. Since cross-entropy loss tries to minimize the probability of other classes as much as possible, the probability of the most likely class will have the largest value. For other classes, the output probabilities will be around zero. It is unfavorable to calculate the per-class decision threshold for detecting unknown intent. In \cite{Hinton2015DistillingTK}, they have proposed temperature scaling to distill knowledge in neural networks by generating soft labels for student network. Here we apply temperature scaling to \textbf{soften} the output (i.e., raises the entropy) of softmax.

\subsubsection{Temperature Scaling}
Given a neural network trained for N-classes classification and an input $\mathbf{x_i}$, the network outputs a class prediction $\hat{y}_i = \argmax_{n} (\mathbf{z}_i)^{(n)} $. The network logits $\mathbf{z_i}$ are vectors, and confidence score $\hat{p}_i$ is computed by softmax function $\sigma_{\rm SM}$ as the following:

\begin{align}
	\sigma_{\rm SM}(\mathbf {z_i})^{(n)} &= \frac {\exp(z_i^{(n)})}{\sum _{j=1}^{N}\exp(z_i^{(j)})} \\
	\hat p_i &= \max_{n} \sigma_{\rm SM}(\mathbf {z_i})^{(n)} 	
\end{align}

We define SofterMax $\hat\sigma_{\rm SM}$ and the soften confidence score $\hat{q}_i$ as the following:
\begin{align}
	\hat\sigma_{\rm SM}(\mathbf {z_i})^{(n)} &= \sigma_{\rm SM}(\mathbf{z_i}/T)^{(n)} \\
	\hat q_i &= \max_{n} \hat\sigma_{\rm SM}(\mathbf {z_i})^{(n)}
\end{align}

where $T$ is the temperature parameter. $\sigma_{\rm SM}$ is a special case of $\hat\sigma_{\rm SM}$ when $T$ equals to 1. We can produce a more conservative probability distribution over classes with $T>1$. When $T$ approaches to infinity, the probability $\hat q_i$ approaches to $\frac{1}{N}$ and degenerate into a uniform distribution, which means that entropy reaches its maximum. 

Choosing a suitable $T$ within a narrow range is essential.\cite{Hinton2015DistillingTK}. $T$ is an empirical hyperparameter that can be tuned in the validation set and it requires unknown examples to do so. Therefore, to get a suitable T, we perform probability calibration on the model by temperature scaling \cite{Guo2017OnCO}. The method does not require unknown examples and get optimal $T$ automatically through optimization. 

\subsubsection{Probability Calibration}
We consider A model as \textbf{well-calibrated} when the calibrated confidence is close to its ground truth correctness likelihood. Our goal is to transform the original confidence $\hat{p}_i$ into the calibrated confidence $\hat q_i$ with temperature scaling. Given a one-hot representation $t$ and a model prediction $y$ , we can express the negative log-likelihood in an example like the following:
\begin{align}
	\mathcal{L} = - \sum_{j=1}^N t_j \log{y_j}
\end{align}
Then we optimize $\hat T$  with respect to negative log-likelihood on the validation set to calibrate the confidence score. We obtain the optimal temperature parameter for SofterMax $\hat\sigma_{\rm SM}$ via probability calibration and set $T$ equal to $\hat T$ during testing. By applying temperature scaling on softmax, SofterMax retains a relatively more conservative probability distribution for all classes as illustrated in Figure \ref{softermax}. Note that we set $T$ equals to 1 during training. Besides, the probability calibration will not affect the prediction result of known intents.

\subsubsection{Decision Boundary}
We further tighten the decision boundary of SofterMax outputs by calculating the probability threshold for each class $c_i$ to detect unknown intents. We first calculate the mean $\mu_i$ and standard deviation $\sigma_i$ of $p(y=c_i|x_j, y_j=c_i)$ for each class where $j$ denotes for the $j^{th}$ examples, so that each class $c_i$ will have their own probability threshold $t_i$. It can further reduce the open space risk in the probability space. We calculate $t_i$ as the following:
\begin{align}
	t_i = \max\{0.5, \mu_i - \alpha \sigma_i\}
\end{align}
The intuition is that we will treat the example whose probability score is $\alpha$ standard deviations away from mean value as an outlier. For instance, if an example's SofterMax output confidence of each class is lower than the corresponding probability threshold, it will be considered as unknown. 

To compare the confidence between different samples, we must calculate a single, comparable confidence score for each sample. We subtract the per-class probability thresholds $t_i$ from the calibrated confidence scores and take the highest value among categories. The lower the confidence score it has, the more likely it is an unknown intent. If the confidence score is lower than 0, we consider the example as unknown. For each sample, we transform the confidence score over classes into a single confidence score as the following:
\begin{align}
	\text{confidence}_{j,i} &= p_{j,i}-t_i \\
	\text{confidence}_j &=\max_i(\text{confidence}_{j,i})
\end{align}
Since Softmax is a non-linear transformation, the logits after temperature scaling are not completely collinear with original logits. Therefore, when the same per-class probability thresholding method is applied to the calibration confidence score, we can obtain different results for unknown intent detection.

In short, we calculate the confidence score of SofterMax by applying temperature scaling to logits, subtracting the per-class probability thresholds, and taking the maximum value of the score in the sample. We will use the confidence score to perform joint prediction in the proposed SMDN method.

\subsection{Deep Novelty Detection}
We further combine novelty detection algorithm with feature representations learned by the deep neural networks to detect unknown intents from different perspectives.

OpenMax \cite{Bendale2016TowardsOS} has demonstrated the potential of open space risk reduction in the feature space. Different from OpenMax which uses logits as the feature space, we use the feature representations from the hidden layer before logits as the feature space. The reason is that previous studies \cite{ouyang2015deepid} have shown that the feature representations contain more high-level semantic concepts than logits. 

Then, we use local outlier factor (LOF) \cite{breunig2000lof}, a density-based novelty detection method, to reduce the open space risk in the feature space and discover unknown intents. It can detect unknown intents in the local context based on local density. We compute LOF as the following:
\begin{align}
	\text{LOF}_k(A) = \frac{\sum_{B \in N_k(A) \frac{lrd(B)}{lrd(A)}}}{|N_k(A)|}
\end{align}
where $N_k(A)$ denotes the set of k-nearest neighbors and \emph{lrd} denotes the local reachability density. We define \emph{lrd} as the following:
\begin{align}
	\text{lrd}_k(A) = \frac{|N_{k}(A)|}{\sum_{B\in N_k(A)}} \text{reachdist}_k(A,B)
\end{align}
which is the reciprocal of the average reachability distance between object A and its neighbors. We define $\text{reachdist}_k(A,B)$ as the following:
\begin{align}
	\text{reachdist}_k(A,B) = \max\{{\text{k-distance}(B), d(A,B)}\}
\end{align}
where d(A,B) denotes the direct distance between A and B, and k-distance denotes the distance of the object A to the $k^{th}$ nearest neighbor. If an example's local density is significantly lower than its k-nearest neighbor's local density, then the example is more likely to be considered as abnormal. We treat the LOF score as the novelty score. The higher the novelty score it has, the more likely it is an unknown intent. Besides, we believe that other novelty detection algorithms are also applicable to this method.

%% Platt scaling
\subsection{SMDN}
Finally, the proposed SMDN method combines the results of SofterMax and deep novelty detection to make the joint prediction. However, the confidence score calculated by SofterMax and the novelty score of LOF cannot be directly combined because they are not on the same scale. Therefore, we use Platt scaling \cite{platt1999probabilistic} to transform the scores into the novelty probability (between 0 and 1) and make the joint prediction. 

Platt scaling can transform scores into probabilities by fitting a logistic regression model to the scores, which is widely used in maximum-margin methods such as SVM.  It can produce probability estimation as the following: 
\begin{align}
	P(y=1|x) = \frac{1}{1+\exp(Af(x)+B)}
\end{align}
where f(x) denotes the scores. A and B is the scaler parameters learned by the algorithm. The goal of Platt scaling is to let the samples near decision boundary have the 50\% chance (novelty probability) to be considered as unknown, and scale the scores of the remaining samples into probability ranged from 0 to 1. After normalization, we can estimate the degree of novelty under the same measurement for SMDN and LOF, and thus make the joint prediction.

\begin{table*}[t!]
\centering
\begin{tabular}{llllllll}
\toprule
  Dataset & Classes & Vocabulary & \#Training & \#Validation & \#Test & Turn-taking & Class distribution \\
  \hline
  SNIPS & 7 & 11,971 & 13,084 & 700 & 700 & Single-turn& Balanced \\
  ATIS & 18 & 938 & 4978 & 500 & 893 & Single-turn& Imbalanced\\
  SwDA & 42 & 21,812 & 162,862 & 20,784 & 20,146 & Multi-turn& Imbalanced\\
\bottomrule
\end{tabular}
\caption{ \label{data-stat-table}  Statistics of ATIS, SNIPS, and SwDA dataset. \# indicates the total number of utterances.}
\end{table*}

\section{Experiments}
\subsection{Datasets}
To investigate the robustness and effectiveness of the proposed methods, we have conducted experiments on three publicly available benchmark dialogue datasets, including SNIPS, ATIS \cite{hemphill1990atis}, and SwDA \cite{Jurafsky1997SwitchboardDL} under different experimental settings. We show the detailed statistics of datasets in Table~\ref{data-stat-table}. 

\textbf{SNIPS \footnote{https://github.com/snipsco/nlu-benchmark/tree/master/2017-06-custom-intent-engines}}
We firstly conduct experiments on SNIPS personal voice assistant dataset. It contains 7 types of user intents across different domains. There are 13084 utterances for training and 700 utterances for validation and testing respectively. The instances in each class are relatively balanced.

\textbf{ATIS (Airline Travel Information System) \footnote{https://github.com/yvchen/JointSLU/tree/master/data}} ATIS dataset contains recordings of people making flight reservations with 18 types of user intent in flight domain. There are 4978 utterances for training, 500 utterances for validation and 893 utterances for testing. The classes in ATIS are highly imbalanced, where the top 25\% classes account for about 93.7\% of the training data.

\textbf{SwDA (Switchboard Dialog Act Corpus) \footnote{https://github.com/cgpotts/swda}} SwDA is a multi-turn dialogue dataset containing 1155 telephone conversations of two people on pre-specified topics with 42 types of dialogue acts. Since the dialogue act can be treated as low-level intent, we would like to know whether the existing detection method is still valid under this scenario. However, there is no standard scheme for splitting SwDA dataset into training, validation and test sets. We follow the data splitting scheme suggested in \cite{Liu2017UsingCI} by randomly sampling 80\% of conversation for the training set, 10\% for the validation set and 10\% for the test set. There are 162862 utterances for training, 20784 utterances for validation and 20146 utterances for testing. Each conversation contains 176 sentences on average. Besides, the classes in SwDA are highly imbalanced, where the top 25\% classes account for about 90.9\% of training data. We will further describe how to split existing datasets with different percentages of known intent Section 4.3.

\subsection{Baseline methods}
We have compared our detection methods with a simple baseline, the state-of-the-art method and its extension, DOC (Softmax).

\begin{enumerate}[noitemsep,topsep=0pt]
  \item \textbf{Softmax (t=0.5)} 
We apply a confidence threshold on the softmax outputs as the simplest baseline where we set the threshold as 0.5. For instance, if the output probability of each class does not exceed 0.5, then the example will be considered as unknown.
  \item \textbf{DOC} \cite{Shu2017DOCDO} is the current state-of-the-art method in the open-world classification problem. The method rejects unknown examples by using the sigmoid activation function as the final layer. It further tightens the decision boundary of the sigmoid function by calculating the confidence threshold for each class through the statistics approach.
  \item \textbf{DOC (Softmax)} A variant method of DOC which replaces the sigmoid activation function with softmax. 
\end{enumerate}
Note that we do not compare with several other baselines mentioned in \cite{Shu2017DOCDO}, since they had shown that the performance of DOC is significantly better. Besides, we evaluate all detection methods under the same classifier for a fair comparison. 

\begin{table*}[t]
\centering
\begin{tabular}{lllll}
\toprule
  Dataset & Model & Acc* & Acc & Macro F${_1}$ \\
  \hline
  SNIPS & BiLSTM \cite{Goo2018SlotGatedMF} & 97  & \textbf{97.43} & 97.47\\
  ATIS & BiLSTM \cite{Wang2018ABB} & 98.99 & 98.66 & 93.99 \\
  SwDA & CNN+CNN \cite{Liu2017UsingCI} & 78.45 & 77.44 & 50.09\\
\bottomrule
\end{tabular}
\caption{ \label{model-stat-table} Classifier performance of using all classes for training on different datasets (\%). * indicates the performance reported in their original paper.}
\end{table*}

\subsection{Experiment settings}
We follow the similar cross-validation setting in \cite{Fei2016BreakingTC, Shu2017DOCDO} by keeping some classes in training as the unknown and mix them back during testing. Then we vary the number of known classes in training set in the range of 25\%, 50\%, and 75\% classes and use all classes for testing. Using 100\% classes for training is the same as a regular intent classification task. To demonstrate our classifier architecture can model the intent well, we report the results of using 100\% classes for training in Table~\ref{model-stat-table}.

To conduct a fair evaluation for the imbalanced dataset, we randomly select known classes by weighted random sampling without replacement in training set for each run of the experiment. Therefore, if there are more examples of intents, it is more likely to be selected as a known class. Meanwhile, the class with fewer examples can still have a chance to be selected with a certain probability.  We treat other classes which are not selected as the known class. Note that we remove examples considered as the unknown in the training and validation set.

\subsection{Evaluation}
We use macro F1-score as the evaluation metric and evaluate the results on all classes, known classes, and the unknown class. Here we mainly focus on the results of the unknown class for detecting unknown intent and report the average result over 10 runs. 

Given a set of classes $C = \{{C_1, C_2, C_3, ... ,C_N}\}$, we compute macro F1-score as the following:
\begin{align}
	F_{1, macro} &=2\frac{\text{recall}_{macro}\times \text{precision}_{macro}}{\text{recall}_{macro} + \text{precision}_{macro}} \\
	\text{precision}_{macro} &=  \frac{ \sum_{i=1}^N\text{precision}_{C_i}}{N} \text{,} \quad \text{recall}_{macro}=  \frac{ \sum_{i=1}^N\text{recall}_{C_i}}{N} \\
	\text{precision}_{C_i} &=\frac{TP_{C_i}}{TP_{C_i}+FP_{C_i}} \text{,}  \quad   \text{recall}_{C_i}=\frac{TP_{C_i}}{TP_{C_i}+FN_{C_i}}
\end{align}
where ${C_i}$ denotes the individual class in the set of classes C.

For probability calibration, we use Expected Calibration Error \cite{ece1983} (ECE) to evaluate the effectiveness of temperature scaling. The main idea is to divide the confidence outputs into equally sized interval K bins and compute the weighted average of the difference between confidence and accuracy over bins. We compute ECE as the following: 
\begin{align}
	ECE = \sum_{i=1}^K P(i)  * | o_i - e_i|
\end{align}
where $P(i)$ denotes the empirical probability of all examples that fall into bin $i$. Here $o_i$ represents true fraction of positive instances in bin $i$ (accuracy) and $e_i$ is the average post-calibrated confidence in bin $i$. The difference between confidence and accuracy represents the confidence gap. The lower the value of ECE, the better the model is calibrated.

\begin{table}[t]
\centering
\begin{tabular}{lllllllllll}
   & SNIPS &   &  & ATIS &   &  & SwDA &   &  \\
\toprule
\% of known intents & 25\% & 50\% & 75\%  & 25\% & 50\% & 75\%  & 25\% & 50\% & 75\% \\
  \hline

Softmax (t=0.5) & - & 6.15 & 8.32 & 8.14 & 15.3 & 17.2 & 19.3 & 18.4 & 8.36 \\
DOC & 72.5 & 67.9 & 63.9 & 61.6 & 63.8 & 37.7 & 25.4 & 19.6 & 7.63 \\
DOC (Softmax) & 72.8 & 65.7 & 61.8 & 63.6 & 63.3 & 39.7 & 23.6 & 18.9 & 7.67 \\
SofterMax & 78.8 & 70.5 & 67.2 & 67.2 & 65.5 & 40.7 & \textbf{28.0} & \textbf{20.7} & 7.51 \\
LOF & 76.0 & 69.4 & 65.8 & 67.3 & 61.8 & 38.9 & 21.1 & 12.7 & 4.50 \\
SMDN & \textbf{79.8} & \textbf{73.0} & \textbf{71.0} & \textbf{71.1} & \textbf{66.6} & \textbf{41.7} & 20.8 & 18.4 & \textbf{8.44} \\
\bottomrule
\end{tabular}
\caption{ \label{result-table}  Macro F1-Score of unknown intent detection on SNIPS, ATIS and SwDA dataset with different proportion of classes treated as known intents. }
\end{table}

\subsection{Hyperparameters} For training classifiers, we initialize the embedding layer through publicly available GloVe \cite{pennington2014glove} pre-trained word vectors (400 thousand words and 300 dimensions) \footnote{http://nlp.stanford.edu/projects/glove/} and further fine-tune the embedding layer through back-propagation. For BiLSTM model, we set the cell output dimension as 128 and dropout rate as 0.5. For both sentence CNN and context CNN in CNN+CNN model, we set the context window size as 3, the kernel size ranging from 1 to 3, and the number of filter maps as 100. We use Adam optimizer with 0.001 learning rate. The maximum training epoch is set as 30 for ATIS and SNIPS and 100 for SwDA. We set the batch size as 128 for SNIPS and ATIS, and 256 for SwDA. We set the fixed input sequence length as their maximum for ATIS and SNIPS; for SwDA, we set the fixed input sequence length as sequence length\lq s mean plus six standard deviations. We implement all models and methods with Keras framework.

For the confidence thresholds of each class in SofterMax and the decision threshold in LOF, we set the $\alpha$ as 2. The intuition is that an example whose confidence score is two standard deviations away from mean will be considered as an outlier (unknown). We set the number of neighbors in LOF as 20 and bin size K in temperature scaling as 15 as suggested by their original implementation.

\begin{figure}[t!]
  \centering  \includegraphics[width=0.99\linewidth ]{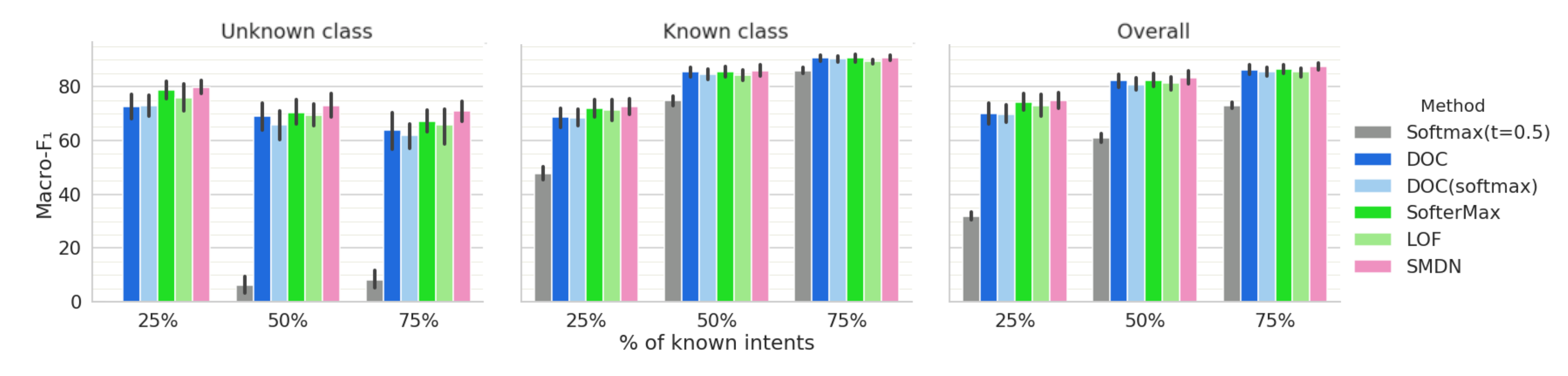}
  \caption{\label{result-SNIPS} Macro F1-Score on SNIPS dataset with different proportion of classes treated as known intents.}
\end{figure}

\begin{figure}[t!]
  \centering  \includegraphics[width=0.99\linewidth ]{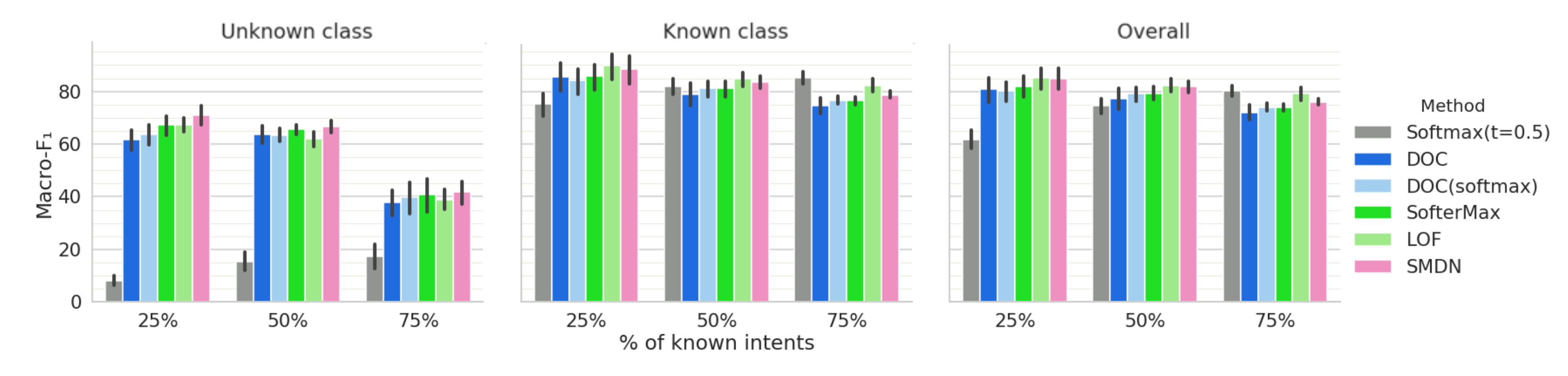}
  \caption{\label{result-ATIS} Macro F1-Score on ATIS dataset with different proportion of classes treated as known intents.}
\end{figure}

\begin{figure}[t!]
  \centering  \includegraphics[width=0.99\linewidth ]{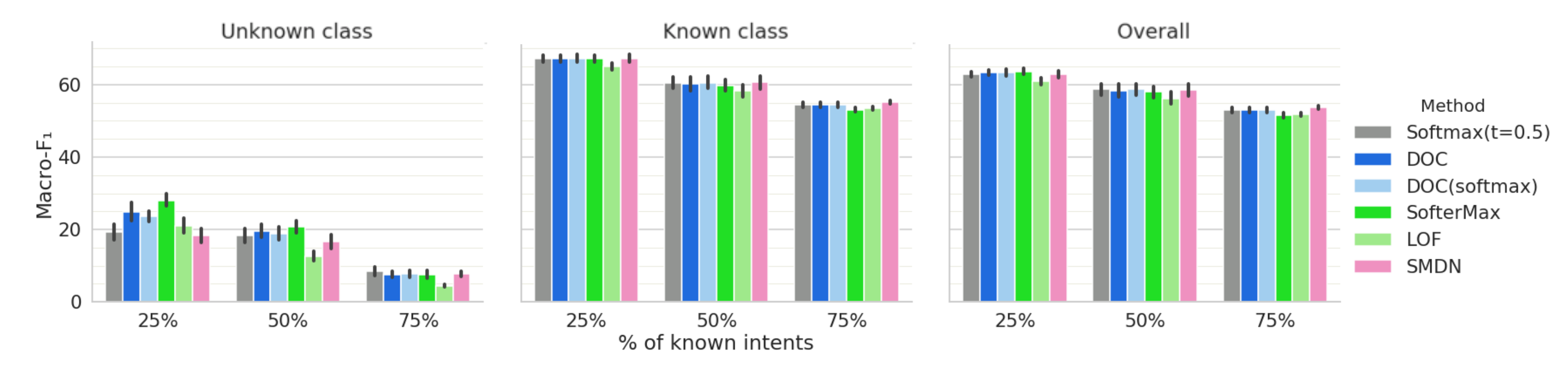}
  \caption{\label{result-SwDA} Macro F1-Score on SwDA dataset with different proportion of classes treated as known intents.}
\end{figure}

\subsection{Results}
In this subsection, we evaluate the results of unknown intent detection on two single-turn dialogue datasets and one multi-turn dialogue dataset. The results are shown in Table \ref{result-table}. Besides, to further investigate how will different detection methods affect the existed classification task, we also evaluate the classification performance on the known classes and all classes. The results are shown in Figure \ref{result-SNIPS}, \ref{result-ATIS}, and \ref{result-SwDA} where the x-axis denotes the different proportion of classes treated as known intents, and the y-axis denotes the macro F1-score. The black error bars in the charts are 95\% confidence level. Figures are best viewed in color. Note that the macro F1-scores for the unknown class in Figures 5, 6 and 7 are the same as in Table 3. Here we compare its performance with known classes classification and overall classification together.

\textbf{Single-turn dialogue datasets} Firstly, we evaluate the results of unknown intent detection on SNIPS and ATIS in Table \ref{result-table}. In comparison, the proposed SMDN method significantly outperforms all other methods in SNIPS and ATIS datasets. Compared with the state-of-the-art baseline method, DOC, SMDN improves SNIPS dataset by 7.3\% in 25\% setting, 5.1\% in 50\% setting and 7.1\% in 75\% setting. Meanwhile, for the proposed SofterMax, it consistently performs better than all baselines in all datasets; for LOF, it performs better than baselines in SNIPS and similar in ATIS.

As for Softmax(t=0.5), it fails to detect unknown intents for both SNIPS and ATIS. There is no notable difference between the results of the DOC and DOC(softmax). It may imply that the key of DOC is the statistic confidence thresholds rather than the 1-vs-rest final layer of the sigmoid. When the number of known intents increases, the error bars in Figure \ref{result-SNIPS} and \ref{result-ATIS}  become smaller, indicating that the results become more robust.

We further evaluate the performance of the overall classification in Figure \ref{result-SNIPS} and \ref{result-ATIS}. As we can see, the proposed SofterMax and SMDN method benefit not only the unknown intent detection but also the overall classification.

\textbf{Multi-turn dialogue dataset} After we verify the effectiveness of the proposed methods on single-turn dialogue datasets, we evaluate the results of unknown intent detection on SwDA dataset. In Table \ref{result-table}, we can see the proposed SofterMax method consistently performs better than all baselines in the 25\%, 50\% settings, and the proposed SMDN method manifests slightly better performance in 75\% setting. For the results of the overall classification shown in Figure \ref{result-SwDA}, the proposed SofterMax method can benefit the unknown intent detection without degrading the overall classification performance.

\begin{table}[t]
\centering
\begin{tabular}{lllllllllll}
   & SNIPS &   &  & ATIS &   &  & SwDA &   &  \\
\toprule
\% of known classes & 25\% & 50\% & 75\%  & 25\% & 50\% & 75\%  & 25\% & 50\% & 75\% \\
  \hline
  Temperature & 1.44 & 1.48 & 1.28 & 1.49 & 1.34 & 1.36 & 1.34 & 1.27 & 1.33 \\
  ECE (Uncalibrated) & 0.01\% & 0.11\% & 0.14\% & 0.01\% & 0.52\% & 0.77\% & 5.26\% & 5.98\% & 7.45\% \\
  ECE (Temp. Scaling) &0.1\% & 0.16\% & 0.14\% & 0.04\% & 0.67\% & 0.85\% & 2.63\% & 2.58\% & 2.86 \% \\
\bottomrule
\end{tabular}
\caption{ \label{temperature-stat-table}  Temperature parameter optimized with respect to negative log-likelihood and expected calibration error (ECE, the lower the better) after temperature scaling. ECE $<$ 1\% represent the model is well-calibrated. }
\end{table}

\begin{figure}[t]
\begin{minipage}{.33\textwidth}
  \centering
  \includegraphics[width= 0.99 \linewidth]{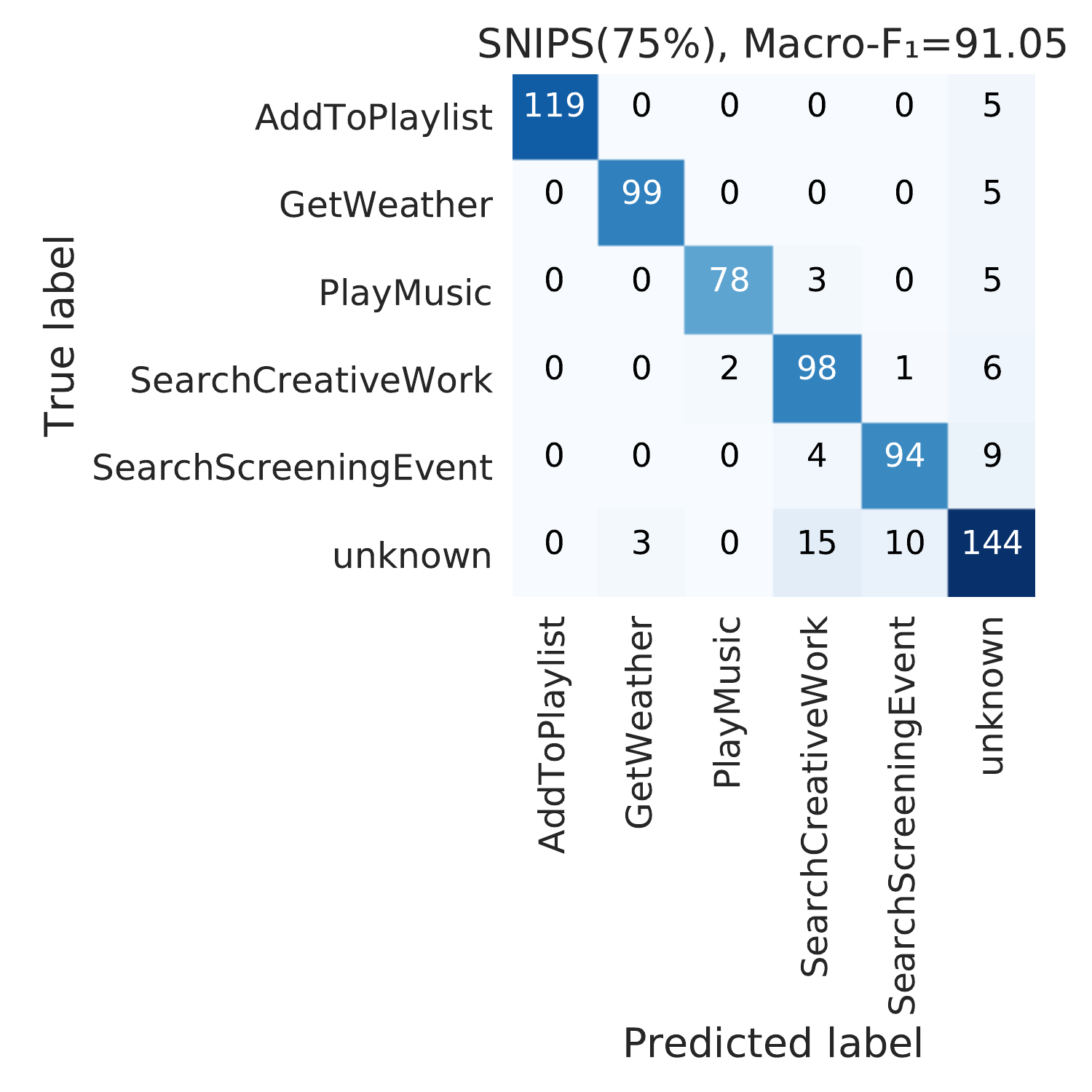}
\end{minipage}
\begin{minipage}{.33\textwidth}
  \centering
  \includegraphics[width= 0.99 \linewidth]{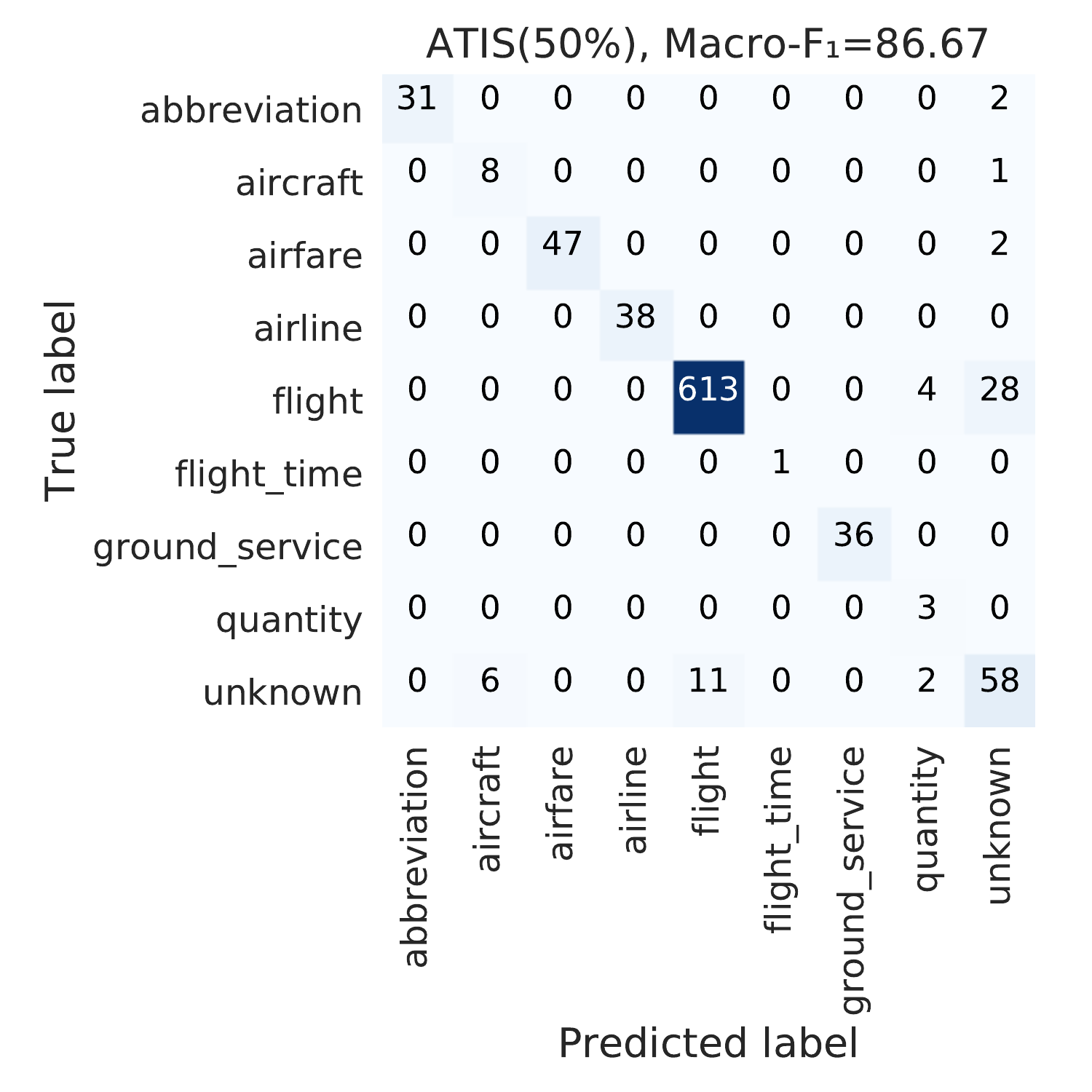}
\end{minipage}
\begin{minipage}{.33\textwidth}
  \centering
  \includegraphics[width=\linewidth]{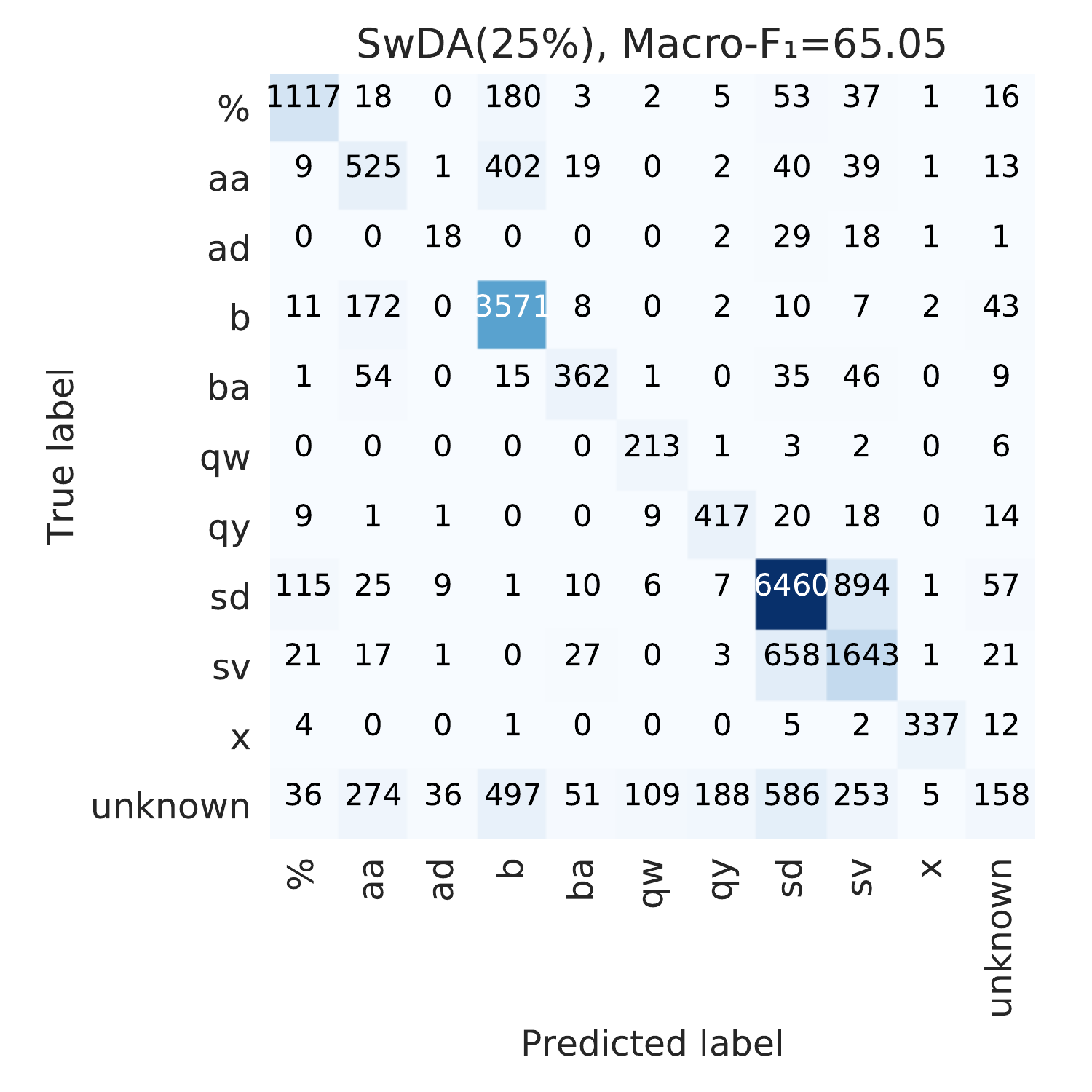}
\end{minipage}
\caption{ \label{mat} Confusion matrix for the overall classification results with SMDN method on three different datasets. The values along the diagonal represent how many examples are correctly classified into the corresponding class. The larger the number, the deeper the color.}
\end{figure}

% Temperature
We also report the temperature parameter and ECE in Table \ref{temperature-stat-table} to evaluate the performance of probability calibration. Temperature parameters that are automatically learned by probability calibration are typically in the range of 1.2 to 1.5. After calibration, the ECE has not changed much in both SNIPS and ATIS datasets, and we reduce the ECE by 3\% to 4\% in SwDA dataset. Notes that we use the validation set which does not contain any unknown intents to calculate ECE. It can only represent how well the model is calibrated with in-domain samples and can not represent the real situation when facing unknown intents. Since we can not merge the probability of unknown intent detection and known intent classification as a whole, we can not calculate the ECE during testing.

Finally, we visualize the experiment results through the confusion matrix in Figure \ref{mat} to demonstrate the effectiveness of SMDN. Results have shown that our method can effectively identify unknown intents under different dialogue scenarios since the classification results are mostly lying on the diagonal.

\section{Discussion}
% Different methods
From the results of different methods, we can observe that merely setting the softmax output confidence threshold as 0.5 does not work well. The DOC method improves performance of unknown intent detection since it calculates the sigmoid output confidence threshold for each known classes based on statistics. However, we can replace the last activation layer of DOC from sigmoid to softmax and still can get a similar performance. Based on the DOC (softmax), SofterMax performs probability calibration on the pre-trained classifier through temperature scaling. By calibrating the confidence of the classifier, we can obtain more reasonable confidence thresholds for each class, hence get better performances than baseline methods. 

\begin{figure}[t]
\begin{minipage}{.49\textwidth}
  \centering
  \includegraphics[width= 0.99 \linewidth]{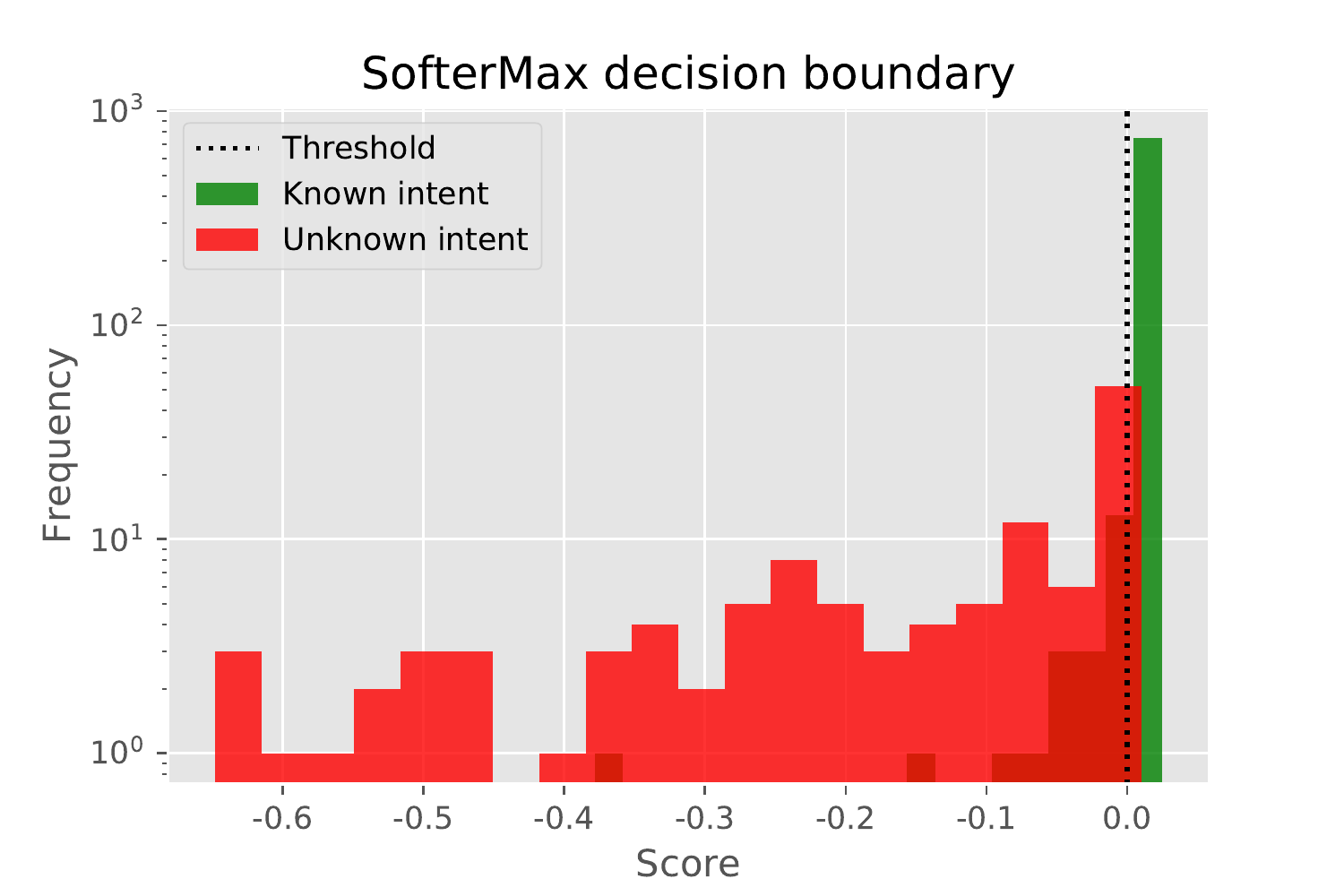}
  \caption{ \label{confidence-score} Confidence score distribution of SofterMax.}
\end{minipage}
\begin{minipage}{.49\textwidth}
  \centering
  \includegraphics[width= 0.99 \linewidth]{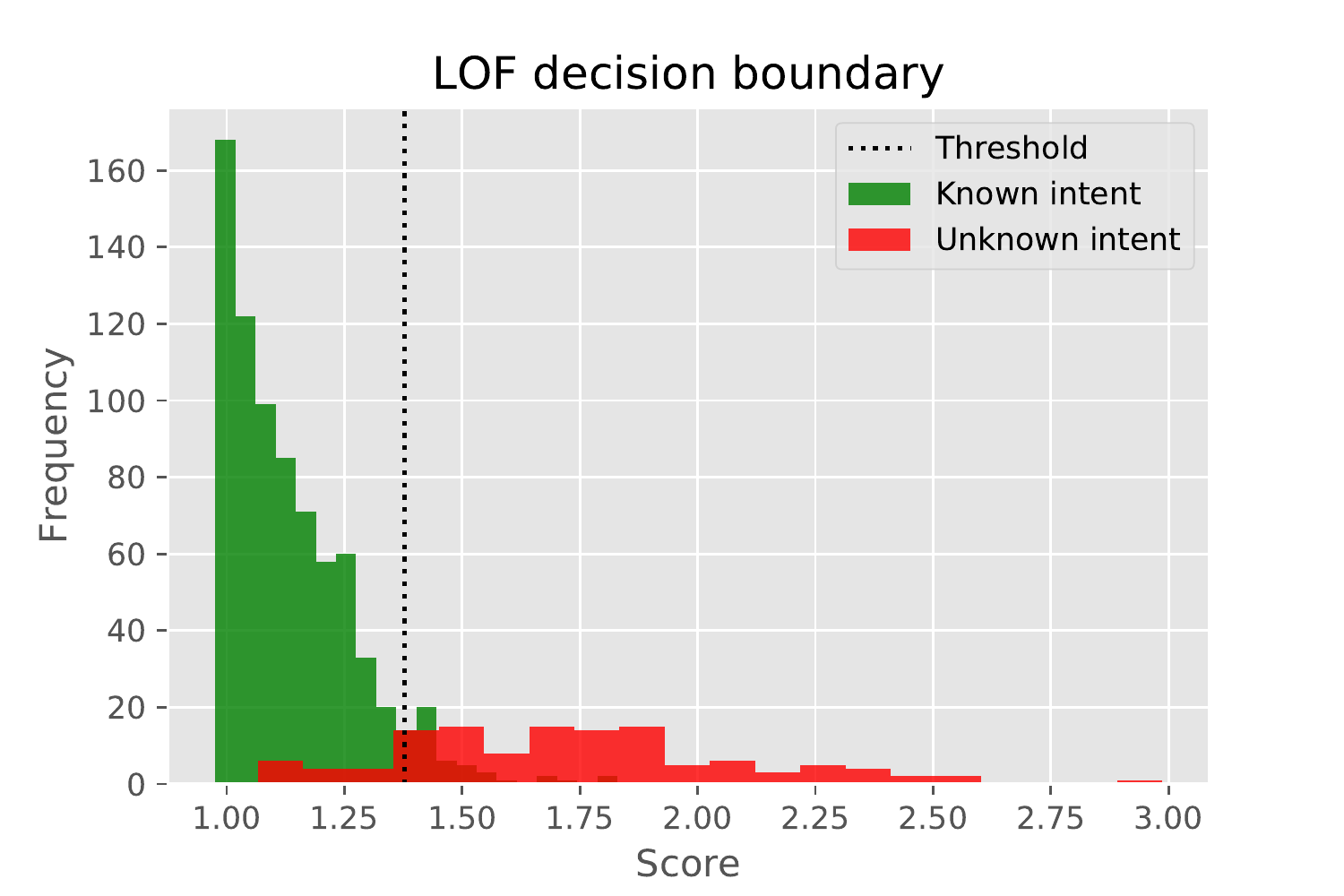}
\caption{ \label{novelty-score} Novelty score distribution of LOF.}
\end{minipage}
\end{figure}

\begin{figure}[t]
\begin{minipage}{.33\textwidth}
  \centering
  \includegraphics[width= 0.99 \linewidth]{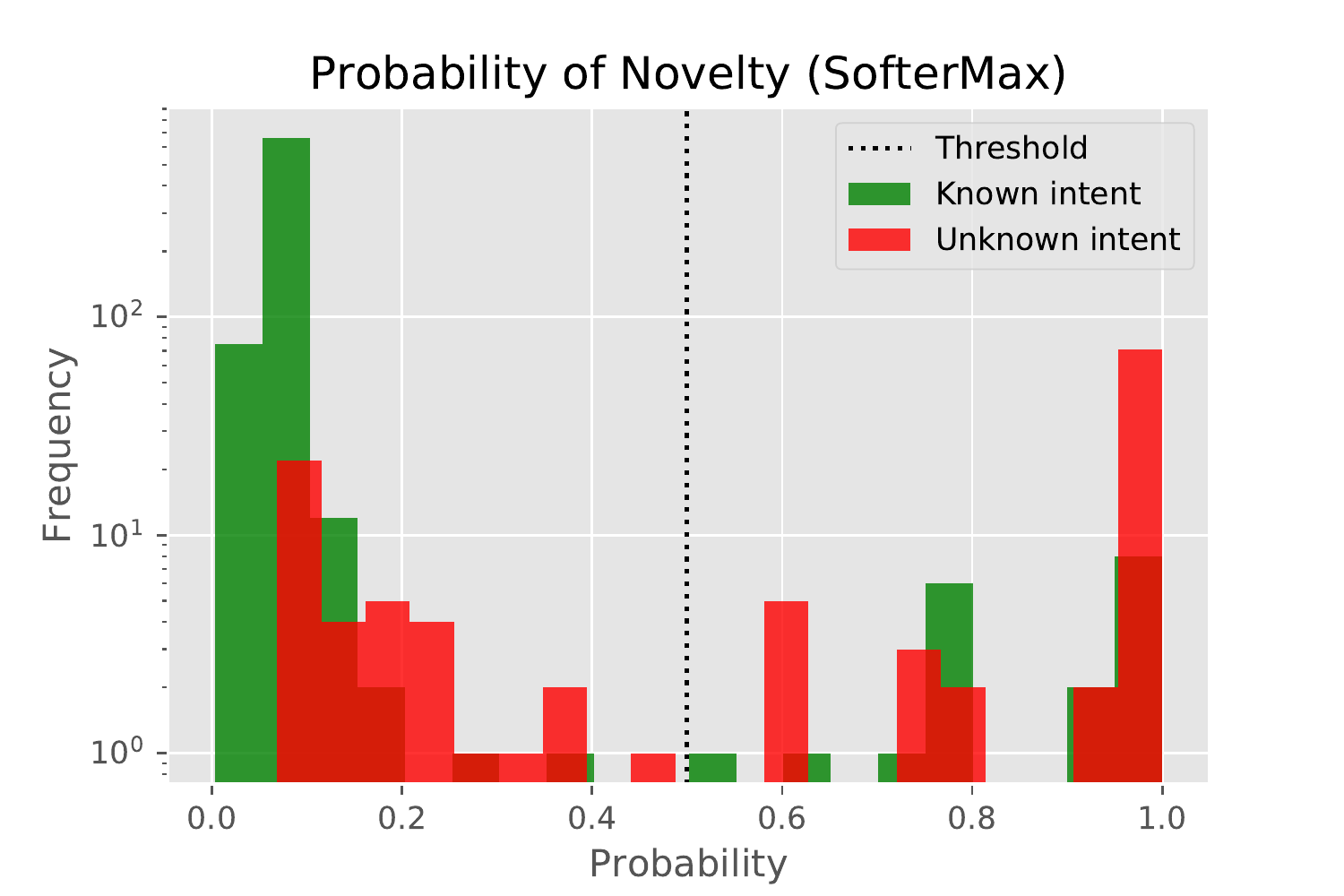}
\end{minipage}
\begin{minipage}{.33\textwidth}
  \centering
  \includegraphics[width= 0.99 \linewidth]{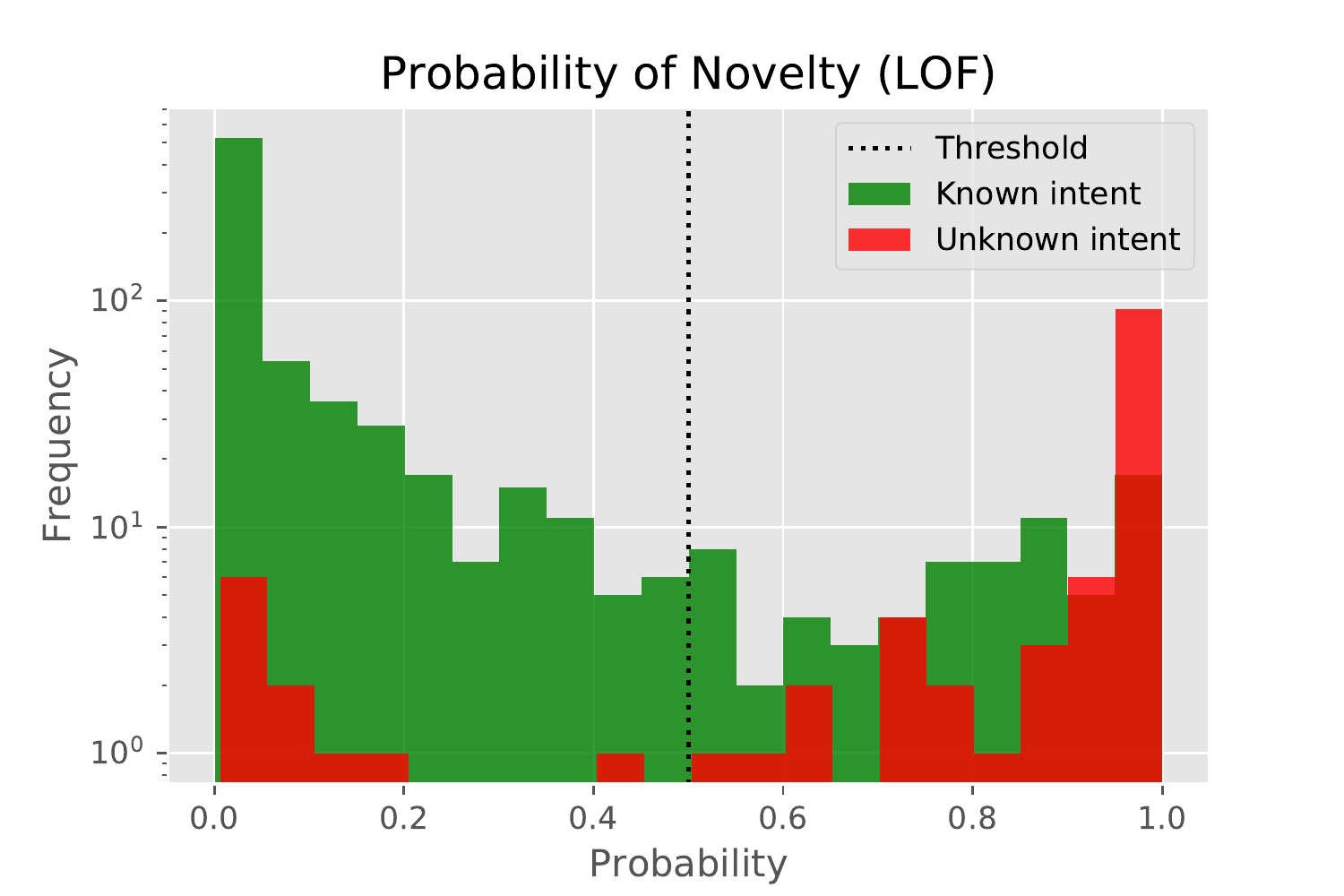}
\end{minipage}
\begin{minipage}{.33\textwidth}
  \centering
  \includegraphics[width= 0.99 \linewidth]{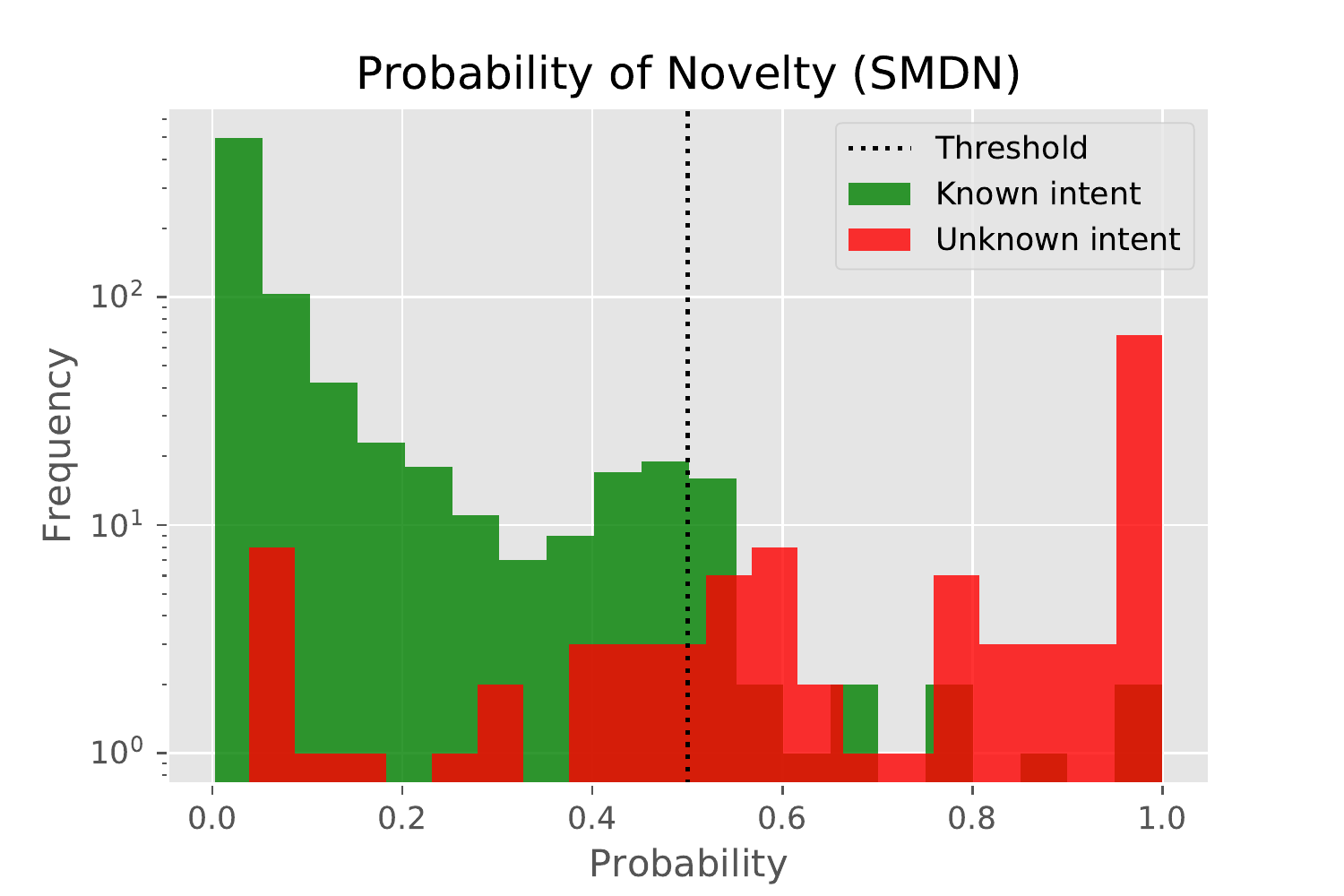}
\end{minipage}
\caption{ \label{novelty-proba} The distribution of novelty probability after Platt scaling for SofterMax, LOF and SMDN.}
\end{figure}

We further detect the unknown intent from different perspectives by feeding feature representations learned by the deep neural network to LOF algorithm. Then, we convert the confidence score of SofterMax and the novelty score of LOF into probabilities through the Platt Scaling. Finally, we obtain the final prediction of SMDN through the joint prediction of SofterMax and LOF. Although the overall performance of LOF is not as good as the SofterMax, we can still greatly improved performance of SMDN through joint prediction.

% Different proportions
By taking the different portion of classes as known intents, we can see that when the number of known intents increases, the performance of almost all methods decreases. Taking the results on SNIPS dataset as an example, the macro F1-score of SMDN method drops from 0.798 to 0.71 when the proportion of known intents increases from 25\% to 75\%. The reason is that when there are more known intents, the semantic meaning of the unknown class is partially overlapping with known classes. Especially in the imbalanced datasets like ATIS and SwDA, their performances drop even more.

% Different datasets
We also observe that the results of SwDA dataset are worse than SNIPS and ATIS datasets. It may be due to the data quality of SwDA itself.  \cite{lee2016sequential} also point out the inter-labeler agreement (accuracy) of SwDA is merely 0.84. The noisy labels prevent the classifier from capturing the high-level semantic concepts of intents, thereby causing the poor performance on detecting unknown intent. While the base classifier can achieve 0.974 and 0.986 accuracy on SNIPS and ATIS, respectively, it can only achieve 0.774 accuracies on SwDA. Still, our method achieves certain improvement in SwDA compared with baselines.

% Vis on score --> probability
Last but not least, we visualize the distribution of confidence and novelty with Figure \ref{confidence-score} and \ref{novelty-score} respectively. The distribution of scaled novelty probability of SofterMax, LOF, and SMDN are shown in Figure \ref{novelty-proba}. Notes that the y-axis is in log scale for better visualization. The green and red bars represent the score distribution of known and unknown examples, respectively. The vertical dot line indicates the decision threshold for unknown intent detection. For SofterMax in \ref{novelty-score}, if the confidence score of an example is lower than the decision threshold, it will be considered as unknown. For LOF, if the novelty score of an example is higher than the decision threshold, it will be considered as unknown. As we can see in Figure \ref{confidence-score} and \ref{novelty-score}, the decision threshold of SofterMax and LOF can separate the unknown intent from known intents. 

For scaled novelty probability distribution in Figure \ref{novelty-proba}, all scores are transformed into the same probability scale ranging from 0 to 1. If the novelty probability of an example is higher than 0.5, it will be considered as unknown. Notes that we only use the per-class probability threshold to calculate the confidence score in Figure 9. Since the novelty probability of SofterMax is derived from the confidence score, it already considers the per-class probability threshold. Therefore, we use 0.5 as the joint prediction threshold, just like the regular binary classification task.  We can see that SofterMax treats some examples of unknown intents as the known, while LOF treats some example of known intents as the unknown. After the joint prediction, the probability distribution becomes more than separable, which significantly improves the performance of unknown intent detection.

\section{Conclusion}
In this paper, we have proposed a simple yet effective post-processing method, SMDN, for detecting unknown intent in the dialogue system. SMDN can be easily applied to a pre-trained deep neural network classifier and requires no changes in the model architecture. The proposed SofterMax calibrates the confidence of softmax output with temperature scaling to reduce the open space risk in probability space and obtains calibrated decision thresholds for detecting unknown intent. We further combine traditional novelty detection algorithm, LOF, with feature representations learned by the deep neural network. We transform the confidence score of SofterMax and novelty score of LOF into novelty probability to make the joint prediction. 

Extensive experiments have been conducted on three benchmark datasets, including two single-turn and one multi-turn dialogue dataset. The results show that our method can yield significant improvements compared with the state-of-the-art baselines. We also believe our method applies to images. For future work, we plan to design a more robust solution that can distinguish unknown intent from known intents even if their semantics meanings are highly similar. Besides, we also plan to use more powerful pre-trained model such as BERT \cite{devlin2018bert} to maximize the benefits of SofterMax.

\section*{Acknowledgments}
This paper is funded by National Natural Science Foundation of China (Grant No: 61673235) and National Key R\&D Program Projects of China (Grant No: 2018YFC1707605). We would like to thank the anonymous reviewers and Yingwai Shiu for their valuable feedback.

%\section*{References}

\bibliography{mybibfile}

\end{document}